\documentclass{article}

  \PassOptionsToPackage{numbers, compress}{natbib}

    \usepackage[final]{neurips_2020}

\usepackage[utf8]{inputenc} 
\usepackage[T1]{fontenc}    
\usepackage{hyperref}       
\usepackage{url}            
\usepackage{booktabs}       
\usepackage{amsfonts}       
\usepackage{nicefrac}       
\usepackage{microtype}      
\usepackage{amsmath}

\usepackage{algorithm}
\usepackage{booktabs} 
\usepackage{subcaption}
\usepackage{caption}
\usepackage{xcolor}
\usepackage[noend]{algpseudocode}
\usepackage{wrapfig}
\usepackage{adjustbox}
\usepackage{xcolor}

\definecolor{forestgreen}{rgb}{0.14, 0.55, 0.14}

\renewcommand{\paragraph}[1]{\vspace{0.20ex}\noindent\textbf{#1}}

\def\eg{\emph{e.g.~}}
\def\etal{{\em et al.~}}
\def\ie{\emph{i.e.~}}

\title{Overparametrization improves robustness against adversarial attacks: A replication study}

\author{Ali Borji \\
Quintic AI, San Francisco, CA \\
\texttt{aliborji@gmail.com} }

\begin{document}

\maketitle

\begin{abstract}

Overparametrization has become a de facto standard in 
machine learning. Despite numerous efforts, our understanding of how and where overparametrization helps model accuracy and robustness is still limited. To this end, here we conduct an empirical investigation to systemically study and replicate previous findings in this area, in particular the study by Madry \etal~\cite{madry2017towards}. Together with this study, our findings support the "universal law of robustness" recently proposed by Bubeck \etal~\cite{bubeck2021law,bubeck2021universal}. We argue that while critical for robust perception, overparametrization may not be enough to achieve full robustness and smarter architectures (\eg the ones implemented by the human visual cortex) seem inevitable.

\end{abstract}

\section{Introduction}

\begin{figure}[t]
    \centering
    \includegraphics[width=.45\textwidth]{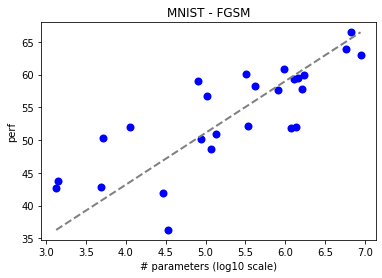}
    \includegraphics[width=.45\textwidth]{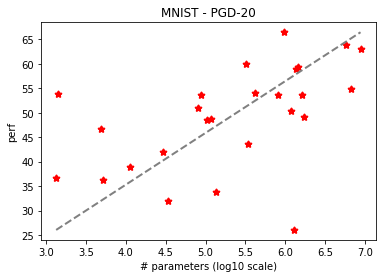} \\
    \includegraphics[width=.45\textwidth]{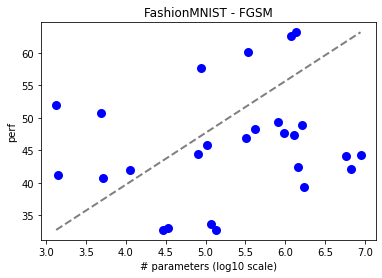}
    \includegraphics[width=.45\textwidth]{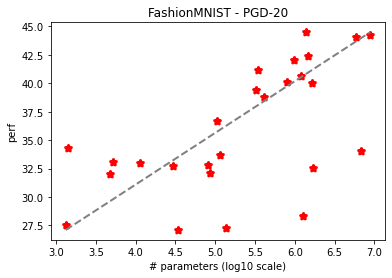} \\
    \includegraphics[width=.45\textwidth]{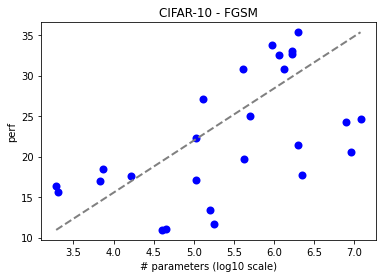}
    \includegraphics[width=.45\textwidth]{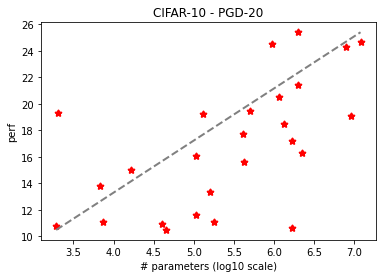} \\
    
    \caption{Classification accuracy (average of standard and robust accuracy against FGSM and PGD-20 attacks) vs. number of model parameters over MNIST, FashionMNIST, and CIFAR-10 datasets. Each dot represents a single model. As can be seen there is a linear relationship between performance and model capacity. The diagonal line represents $y=x$ and is only for the illustration purpose.}
    \label{fig:SUM}
\end{figure}

\begin{figure}[t]
    \centering
    \includegraphics[width=.45\textwidth]{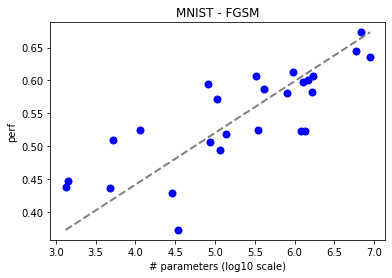}
    \includegraphics[width=.45\textwidth]{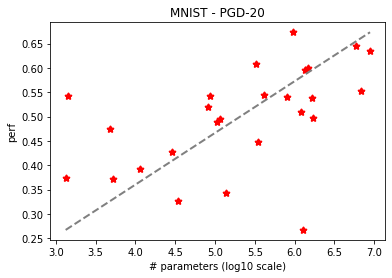} \\
    \includegraphics[width=.45\textwidth]{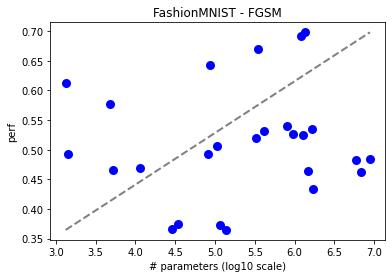}
    \includegraphics[width=.45\textwidth]{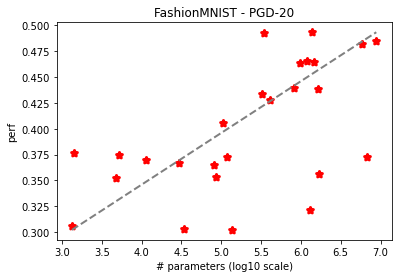} \\
    \includegraphics[width=.45\textwidth]{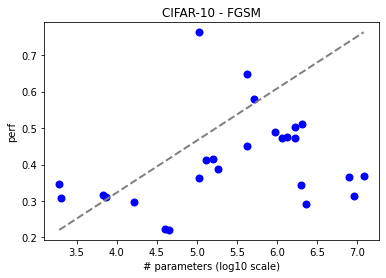}
    \includegraphics[width=.45\textwidth]{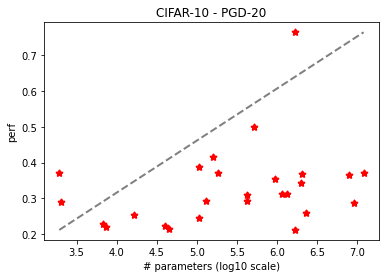} \\
    
    \caption{Classification accuracy (average of standard and robust accuracy) vs. number of model parameters when performance is normalized/divided to standard accuracy for each model. The linear relationship still exists although less pronounced compared to the unnormalized setting shown in Fig.~\ref{fig:SUM}.}
    
    \label{fig:NORMALIZED_SUM}
\end{figure}

Despite being highly overparametrized\footnote{\ie the number of free parameters in their architectures is often several orders of magnitude greater than the number of training examples, yet models do not overfit and perform well on the unseen test data.}, deep neural networks perform very well on a variety of tasks. They are even capable of perfectly fitting randomly labeled data~\cite{zhang2021understanding}. 
Several theoretical works such as \cite{belkin2019reconciling,nakkiran2021deep,poggio2018theory,neyshabur2018towards,
hassani2022curse,tripuraneni2021overparameterization} have studied this phenomenon. Few empirical works, however, have investigated the impact of overparametrization on model robustness against adversarial attacks. Some researchers have experimentally studied the relationship between network capacity/scale and general robustness (\eg~\cite{novak2018sensitivity, xie2019intriguing, gowal2020uncovering}). Following~\cite{madry2017towards}, here we explore the impact of network capacity on standard and robust accuracy systematically over a wider range of networks over MNIST, FashionMNIST and CIFAR-10 datasets. Our results support the findings reported by Madry \etal~\cite{madry2017towards} and show that network capacity impacts performance and robustness.

Recently, Bubeck \etal~\cite{bubeck2021law,bubeck2021universal} have proposed a theory so-called "the universal law of robustness" suggesting that for a model to be robust it has to interpolate the training data smoothly. They prove that for a broad class of data distributions and model classes, overparametrization is necessary if one wants to interpolate the data smoothly. 
Further, they quantitatively determine how many parameters are needed to learn such robust models. This theory also makes a number of predictions for building robust models on larger datasets such as ImageNet. Our work here is an attempt to verify this theory over small datasets.

\section{Experiments and Results}

We trained several variants of 5 different types of CNNs (26 models in total) over three datasets including MNIST, FashionMNIST, and CIFAR-10. Each model was instantiated from a two layer CNN (two conv layers with Relu activation each followed by pooling layer, see appendix A). These networks differ in terms of the number of convolutional and pooling filters, filter sizes, number of neurons in the fully connected layer and dilation size. The complete list of these architectures is given in appendix A. The number of parameters in these networks vary from small (around $10^4$) to large (around $10^6$).

\begin{figure}[t]
    \centering
    \includegraphics[width=.45\textwidth]{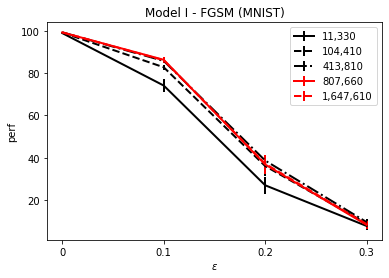}
    \includegraphics[width=.45\textwidth]{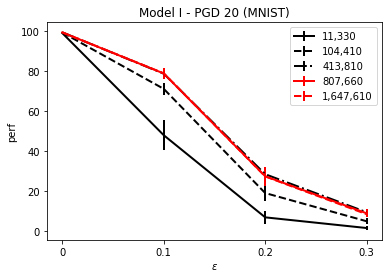} \\
    \includegraphics[width=.45\textwidth]{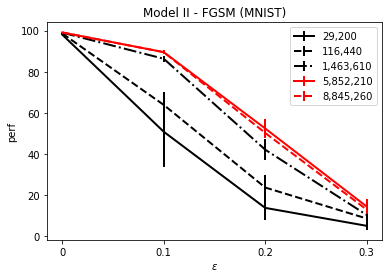}
    \includegraphics[width=.45\textwidth]{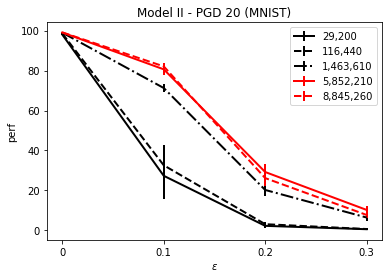} \\
    \includegraphics[width=.45\textwidth]{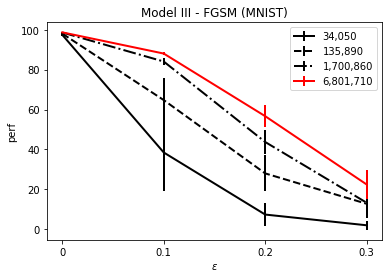}
    \includegraphics[width=.45\textwidth]{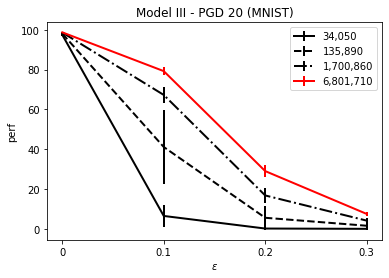} \\
    \includegraphics[width=.45\textwidth]{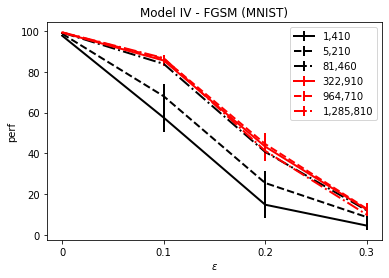}
    \includegraphics[width=.45\textwidth]{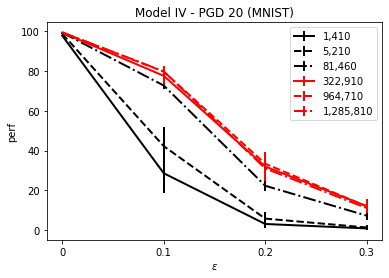} \\
    \includegraphics[width=.45\textwidth]{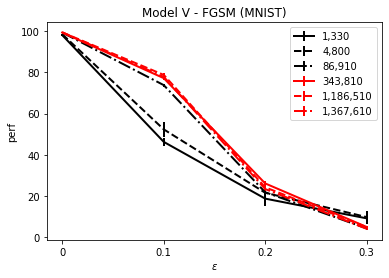}
    \includegraphics[width=.45\textwidth]{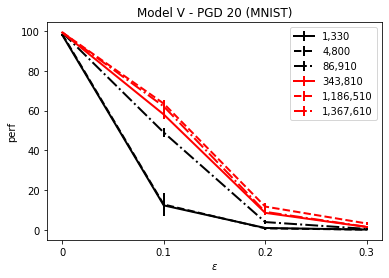}
    \caption{Standard accuracy ($\epsilon=0$) and robust accuracy ($\epsilon > 0$) of models (each row corresponds to one model family) over MNIST dataset for the unnormalized setting corresponding to Fig.~\ref{fig:SUM}. In almost all cases, increasing model capacity improves the accuracy up to a saturation point. The legend shows the number of model parameters.}
    \label{fig:mnist_orig}
\end{figure}

\begin{figure}[t]
    \centering
    \includegraphics[width=.45\textwidth]{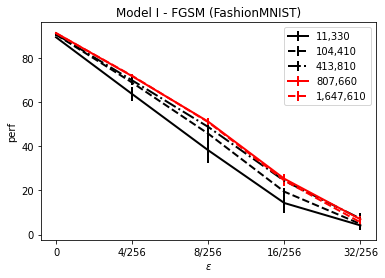}
    \includegraphics[width=.45\textwidth]{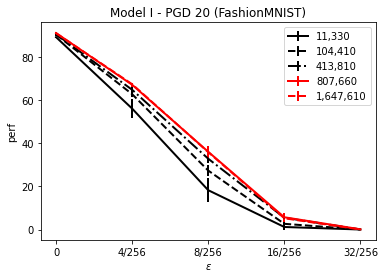} \\
    \includegraphics[width=.45\textwidth]{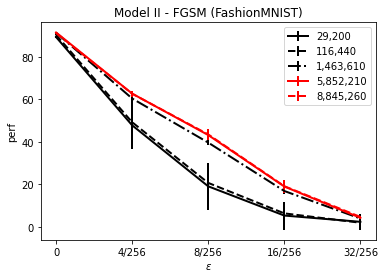}
    \includegraphics[width=.45\textwidth]{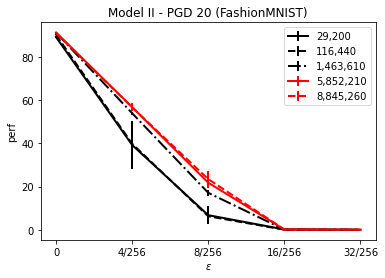} \\
    \includegraphics[width=.45\textwidth]{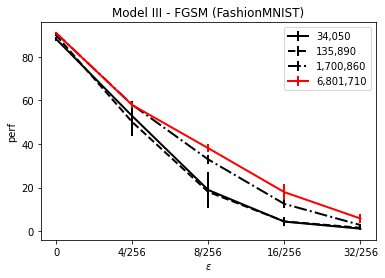}
    \includegraphics[width=.45\textwidth]{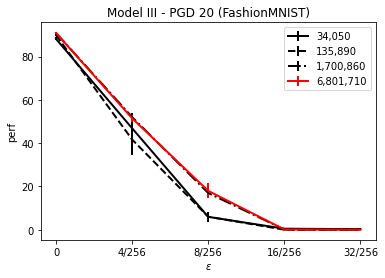} \\
    \includegraphics[width=.45\textwidth]{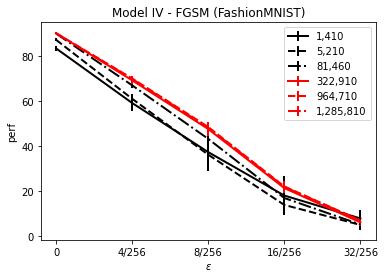}
    \includegraphics[width=.45\textwidth]{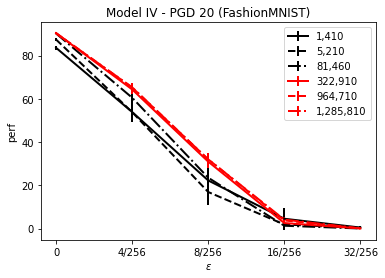} \\
    \includegraphics[width=.45\textwidth]{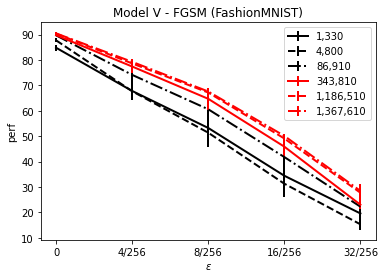}
    \includegraphics[width=.45\textwidth]{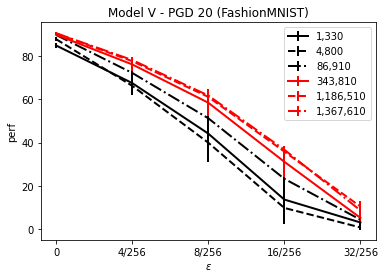}
    \caption{Standard accuracy ($\epsilon=0$) and robust accuracy ($\epsilon > 0$) of models over FashionMNIST dataset for the unnormalized setting corresponding to Fig.~\ref{fig:SUM}.}
    \label{fig:fmnist_orig}
\end{figure}

\begin{figure}[t]
    \centering
    \includegraphics[width=.45\textwidth]{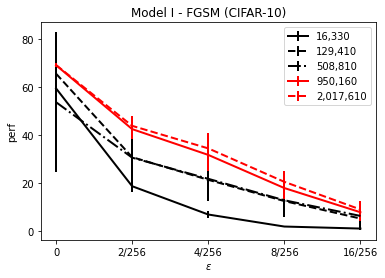}
    \includegraphics[width=.45\textwidth]{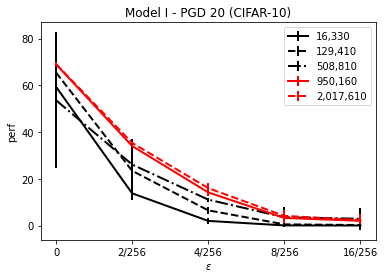} \\
    \includegraphics[width=.45\textwidth]{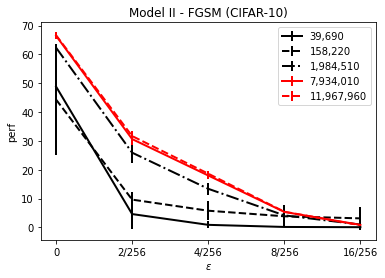}
    \includegraphics[width=.45\textwidth]{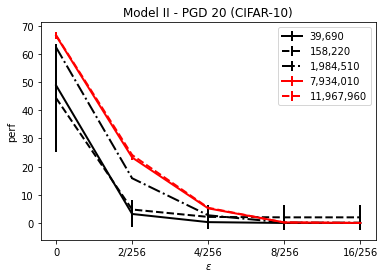} \\
    \includegraphics[width=.45\textwidth]{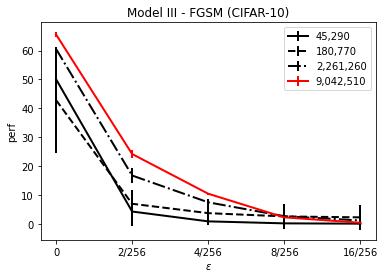}
    \includegraphics[width=.45\textwidth]{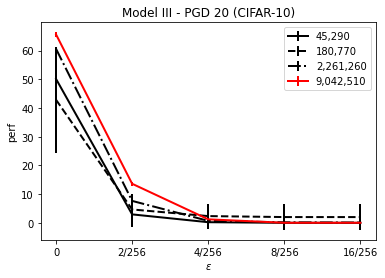} \\
    \includegraphics[width=.45\textwidth]{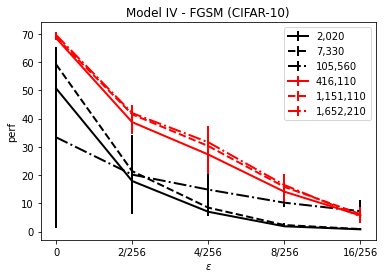}
    \includegraphics[width=.45\textwidth]{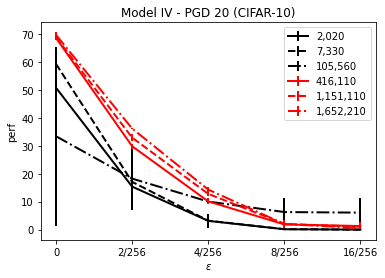} \\
    \includegraphics[width=.45\textwidth]{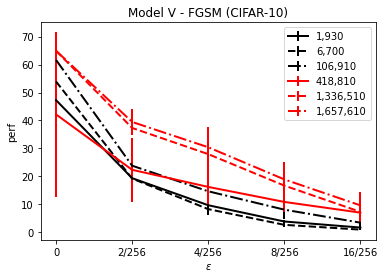}
    \includegraphics[width=.45\textwidth]{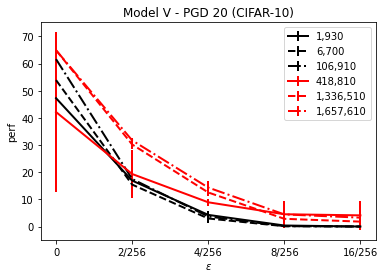}
    \caption{Standard accuracy ($\epsilon=0$) and robust accuracy ($\epsilon > 0$) of models over CIFAR-10 dataset for the unnormalized setting corresponding to Fig.~\ref{fig:SUM}.}
    \label{fig:cifar_orig}
\end{figure}

Each model was trained 5 times (10 epochs over MNIST and FashionMNIST, and 15 epochs over CIFAR-10). Results were then averaged over the runs.
Standard accuracy (\ie original test data with no perturbation) and robust accuracy (against FGSM~\cite{goodfellow2014explaining} and $\emph{l}_\infty$ PGD-20 ($\alpha=\nicefrac{2}{255}$)~\cite{madry2017towards} attacks) were then computed. The average accuracy (standard plus robust) as a function of model parameters is plotted for each of the models in Fig.~\ref{fig:SUM}. As can be seen, there is a linear relationship between model capacity and model performance over the three datasets. The linear correlation is stronger against the PGD attack. Notice that there are some overparametrized models that are less robust than smaller models (and vice versa). Results for individual models are shown in Figs.~\ref{fig:mnist_orig},~\ref{fig:fmnist_orig}, and~\ref{fig:cifar_orig}. Notice that models exhibit enhanced robustness to adversarial examples as number of parameters increases but the advantage saturates at some point.

\begin{figure}[t]
    \centering
    \includegraphics[width=1\textwidth]{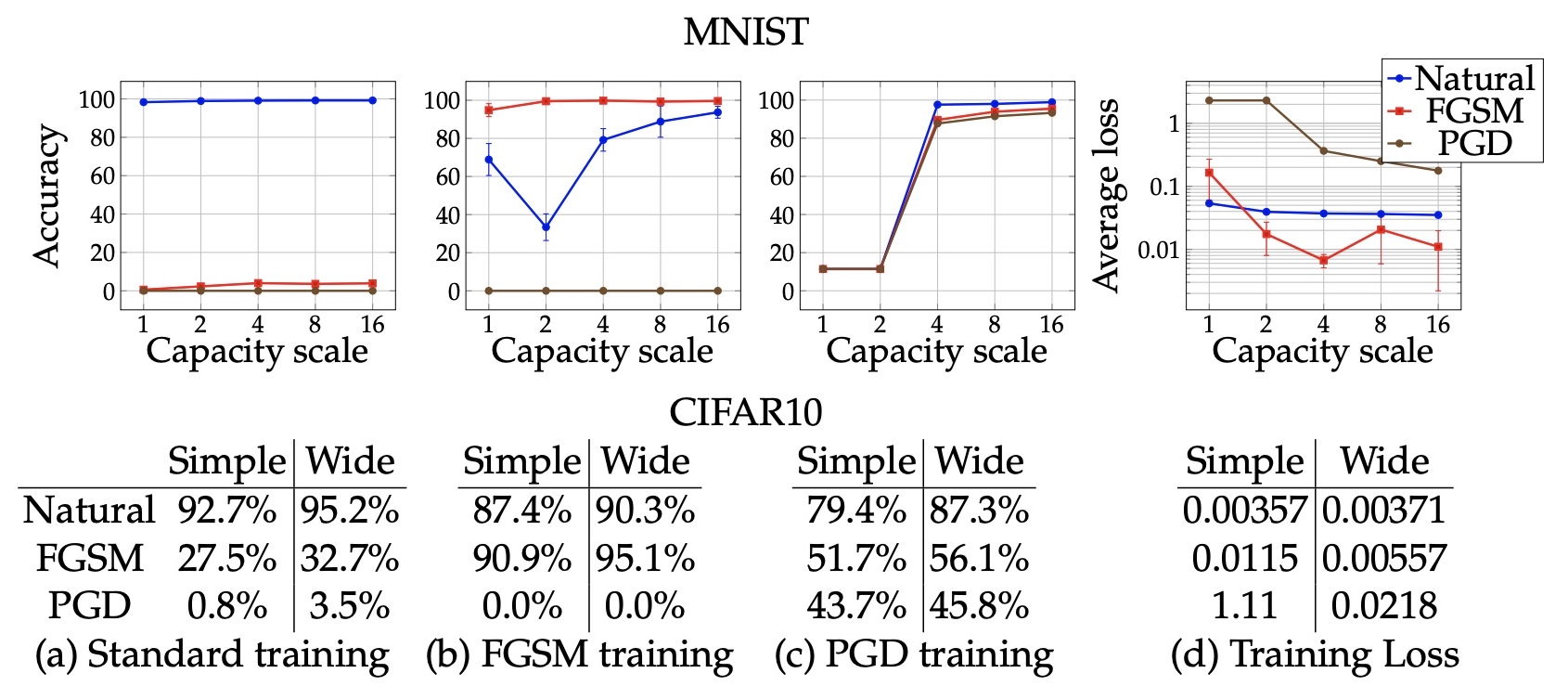}
    \caption{Madry \etal trained networks of varying capacity on MNIST and CIFAR-10 datasets (two layer CNNs over the former two datasets and ResNet over the latter): (a) natural examples, (b) with FGSM-made adversarial examples, (c) with PGD-made adversarial examples. The first three plots/tables show how the standard and adversarial accuracy changes with respect to capacity for each training regime. As can be seen increasing network capacity improves model performance and robustness. Figure from Madry \etal~\cite{madry2017towards}.} 
    \label{fig:madry}
\end{figure}

Overall, our results qualitatively agree with the results reported by Madry~\etal~\cite{madry2017towards} shown in Fig.~\ref{fig:madry}. While here we only considered standard training, they also showed that overparametrized models become more robust with adversarial training compared to smaller models.

One might argue that higher robustness is due to the higher performance of the bigger models in the first place. In other words, larger models are more robust just because they have higher standard accuracy. To address this, we normalized the curves in Figs.~\ref{fig:mnist_orig},~\ref{fig:fmnist_orig}, and~\ref{fig:cifar_orig} to the standard accuracy of each model. Normalized curves for models are shown in Appendix B. 
The normalized perf vs. parameters plots for models is shown in Fig.~\ref{fig:NORMALIZED_SUM}. The linear relationship between network capacity and performance still persists although less pronounced compared to the unnormalized case shown in Fig.~\ref{fig:SUM}.

\section{Discussion and Conclusion}

Our analysis here supports the findings by Madry \etal~\cite{madry2017towards} and Bubeck \etal~\cite{bubeck2021law,bubeck2021universal} with the caveat that Bubeck \etal's analysis is based $\emph{l}_2$ norm whereas attacks considered here (and also by Madry \etal) are based on $\emph{l}_\infty$. This remains to be explored in future work.

An important open problem that should be investigated in future is 
delineating standard and robust accuracy for highly overparametrized models. In other words, 
is higher robustness of larger models due to their higher standard accuracy or these models have inherent properties that makes them more robust. Our preliminary investigation here over normalized models shows that overparametrization offers additional benefits rather than just improving the standard accuracy. Notice that computing robust accuracy is non-trivial (see for example~\cite{recht2019imagenet,shankar2020evaluating}).

Another future direction is the extent to which overparametrization impacts different types of generalizations (\eg generalization to affine transformations such as (in-plane and in-depth) rotation and scale, natural perturbations (\eg blur, noise), out of distribution generalization, as well as additional adversarial attacks).

Here we considered simple models\footnote{Madry \etal also considered ResNet over CIFAR-10} and datasets. Testing the universal law of robustness over larger datasets such as ImageNet is a fantastic future direction. Bubeck \etal predict that robust models on this dataset require number of parameters in order of 10 billions. To put this in perspective, the best available models have around 500 million parameters. Training models at this scale bring along new challenges that should be tackled.

Considering our results and findings from the literature, we believe that while being critical overparametrization is only one side of the coin. The other side of the coin involves more intelligent architectures for robust perception that we have not yet discovered (\eg similar the ones implemented by the brain). For example, it is well known that deep neural networks are biased towards texture and do not rely much on shape~\cite{geirhos2018imagt,borji2020shape}. It seems unlikely that just mere overparametrization can remedy this problem\footnote{Consider a highly overparametrized model trained on original MNIST images. Such a model will not generalize well to negative MNIST images. Humans have no trouble recognizing such digits! Early visual cortex incorporates a complicated circuitry of neurons to preprocess the visual signal before relaying it to higher processing centers.}.

\bibliographystyle{plain}
\bibliography{refs}

\clearpage

\appendix
\section{Model architectures}

\begin{verbatim}
class NetTest(nn.Module):
    def __init__(self, im_res=28, conv1_size=3, conv2_size=3, pool1_size=2, pool2_size=2, 
    num_conv1=10, num_conv2=10, stride_1=1, stride_2=1, padd_1=0, padd_2=0, num_hid=50, 
    dilation_1=1, bias_flag=False):
        super(NetTest, self).__init__()
        self.conv1 = nn.Conv2d(1, num_conv1, kernel_size=conv1_size, stride=stride_1, 
        padding=padd_1, dilation=dilation_1, bias=bias_flag)
        self.res_conv1 = int((im_res - dilation_1*(conv1_size-1)-1 + 2*padd_1) / stride_1+1)
        self.res_pool1 = int((self.res_conv1 - pool1_size + 2*0) / pool1_size + 1)                

        self.conv2 = nn.Conv2d(num_conv1, num_conv2, kernel_size=conv2_size, 
        stride=stride_2, padding=padd_2, bias=bias_flag)
        
        self.res_conv2 = int((self.res_pool1 - conv2_size + 2*padd_2) / stride_2 + 1 )       
        self.res_pool2 = int((self.res_conv2 - pool2_size + 2*0) / pool2_size + 1)               
        
        self.conv2_drop = nn.Dropout2d()

        self.fc1 = nn.Linear((self.res_pool2**2)*num_conv2, num_hid, bias=bias_flag)        
        self.fc2 = nn.Linear(num_hid, 10, bias=bias_flag)
        self.pool1_size = pool1_size
        self.pool2_size = pool2_size
        self.num_conv2 = num_conv2
        self.im_res = im_res


        
    def forward(self, x):
        tmp1 = F.relu(F.max_pool2d(self.conv1(x), self.pool1_size))
        tmp = F.relu(F.max_pool2d(self.conv2_drop(self.conv2(tmp1)), self.pool2_size))

        x = tmp.view(-1, (self.res_pool2**2)*self.num_conv2)
        x = F.relu(self.fc1(x))
        x = F.dropout(x, training=self.training)
        x = self.fc2(x)
        return x   
\end{verbatim}

\begin{verbatim}
    
    
    
    

  # Model I 
  NetTest(conv1_size=5, conv2_size=5, pool1_size=2, pool2_size=2, num_conv1=10, 
  num_conv2=10, num_hid=50, bias_flag=True)
  NetTest(conv1_size=5, conv2_size=5, pool1_size=2, pool2_size=2, num_conv1=50, 
  num_conv2=50, num_hid=50, bias_flag=True)
  NetTest(conv1_size=5, conv2_size=5, pool1_size=2, pool2_size=2, num_conv1=100,
  num_conv2=100, num_hid=100, bias_flag=True)  
  NetTest(conv1_size=5, conv2_size=5, pool1_size=2, pool2_size=2, num_conv1=150, 
  num_conv2=150, num_hid=100, bias_flag=True)
  NetTest(conv1_size=5, conv2_size=5, pool1_size=2, pool2_size=2, num_conv1=200, 
  num_conv2=200, num_hid=200, bias_flag=True)


  # Model II
  NetTest(conv1_size=3, conv2_size=3, pool1_size=1, pool2_size=1, num_conv1=5, 
  num_conv2=5, num_hid=10, bias_flag=True)
  NetTest(conv1_size=3, conv2_size=3, pool1_size=1, pool2_size=1, num_conv1=10, 
  num_conv2=10, num_hid=20, bias_flag=True)
  NetTest(conv1_size=3, conv2_size=3, pool1_size=1, pool2_size=1, num_conv1=50, 
  num_conv2=50, num_hid=50, bias_flag=True)
  NetTest(conv1_size=3, conv2_size=3, pool1_size=1, pool2_size=1, num_conv1=100, 
  num_conv2=100, num_hid=100, bias_flag=True)  
  NetTest(conv1_size=3, conv2_size=3, pool1_size=1, pool2_size=1, num_conv1=150, 
  num_conv2=150, num_hid=100, bias_flag=True)


  # Model III
  NetTest(conv1_size=2, conv2_size=2, pool1_size=1, pool2_size=1, num_conv1=5, 
  num_conv2=5, num_hid=10, bias_flag=True)
  NetTest(conv1_size=2, conv2_size=2, pool1_size=1, pool2_size=1, num_conv1=10, 
  num_conv2=10, num_hid=20, bias_flag=True)
  NetTest(conv1_size=2, conv2_size=2, pool1_size=1, pool2_size=1, num_conv1=50, 
  num_conv2=50,num_hid=50, bias_flag=True)
  NetTest(conv1_size=2, conv2_size=2, pool1_size=1, pool2_size=1, num_conv1=100, 
  num_conv2=100, num_hid=100, bias_flag=True)  


  # Model IV
  NetTest(conv1_size=4, conv2_size=4, pool1_size=2, pool2_size=2, num_conv1=5, 
  num_conv2=5, num_hid=10, bias_flag=True)
  NetTest(conv1_size=4, conv2_size=4, pool1_size=2, pool2_size=2, num_conv1=10, 
  num_conv2=10, num_hid=20, bias_flag=True)
  NetTest(conv1_size=4, conv2_size=4, pool1_size=2, pool2_size=2, num_conv1=50, 
  num_conv2=50, num_hid=50, bias_flag=True)
  NetTest(conv1_size=4, conv2_size=4, pool1_size=2, pool2_size=2, num_conv1=100, 
  num_conv2=100, num_hid=100, bias_flag=True)  
  NetTest(conv1_size=4, conv2_size=4, pool1_size=2, pool2_size=2, num_conv1=200, 
  num_conv2=200, num_hid=100, bias_flag=True)  
  NetTest(conv1_size=4, conv2_size=4, pool1_size=2, pool2_size=2, num_conv1=200, 
  num_conv2=200, num_hid=200, bias_flag=True)  


  # Model V
  NetTest(conv1_size=5, conv2_size=5, pool1_size=2, pool2_size=2, num_conv1=5, 
  num_conv2=5, num_hid=10, dilation_1=2, bias_flag=True).to(device),
  NetTest(conv1_size=5, conv2_size=5, pool1_size=2, pool2_size=2, num_conv1=10, 
  num_conv2=10, num_hid=20, dilation_1=2, bias_flag=True).to(device),
  NetTest(conv1_size=5, conv2_size=5, pool1_size=2, pool2_size=2, num_conv1=50, 
  num_conv2=50, num_hid=50, dilation_1=2, bias_flag=True).to(device),
  NetTest(conv1_size=5, conv2_size=5, pool1_size=2, pool2_size=2, num_conv1=100, 
  num_conv2=100, num_hid=100, dilation_1=2, bias_flag=True).to(device),  
  NetTest(conv1_size=5, conv2_size=5, pool1_size=2, pool2_size=2, num_conv1=200, 
  num_conv2=200, num_hid=100, dilation_1=2, bias_flag=True).to(device),  
  NetTest(conv1_size=5, conv2_size=5, pool1_size=2, pool2_size=2, num_conv1=200, 
  num_conv2=200, num_hid=200, dilation_1=2, bias_flag=True).to(device) ]  
  
  
\end{verbatim}

\clearpage

\section{Results for the unnormalized setting}
The following figures show the results in the unnormalized case. Here, the classification accuracies are divided by the standard accuracy (hence standard accuracy = 1). 

\begin{figure}[t]
    \centering
    \includegraphics[width=.45\textwidth]{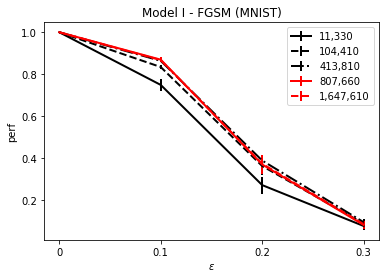}
    \includegraphics[width=.45\textwidth]{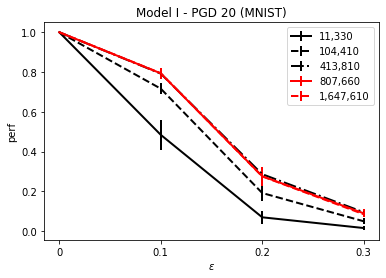} \\
    \includegraphics[width=.45\textwidth]{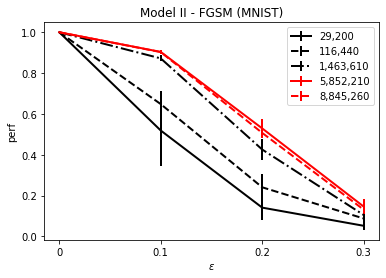}
    \includegraphics[width=.45\textwidth]{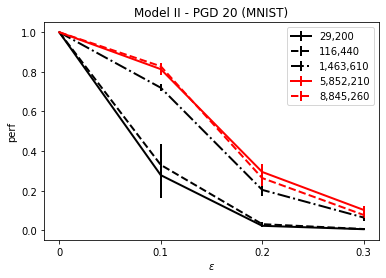} \\
    \includegraphics[width=.45\textwidth]{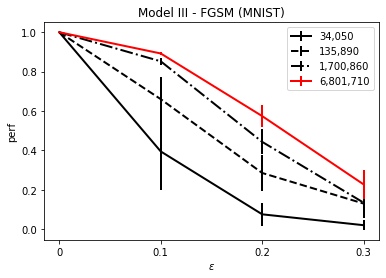}
    \includegraphics[width=.45\textwidth]{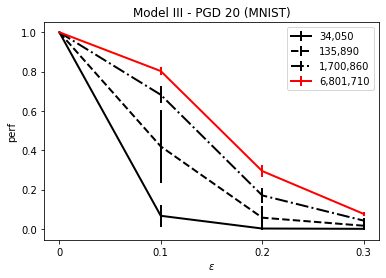} \\
    \includegraphics[width=.45\textwidth]{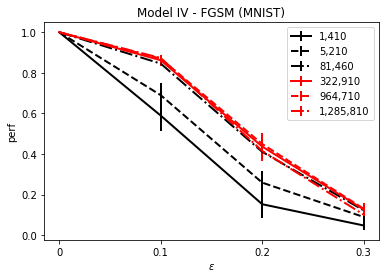}
    \includegraphics[width=.45\textwidth]{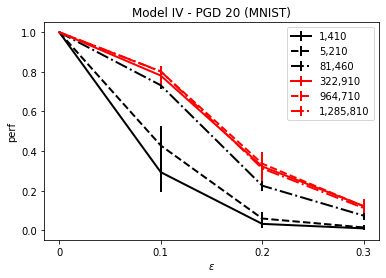} \\
    \includegraphics[width=.45\textwidth]{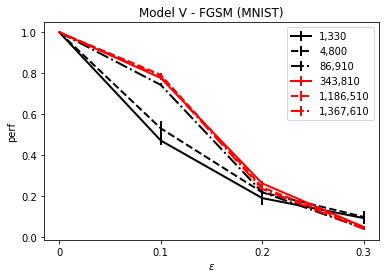}
    \includegraphics[width=.45\textwidth]{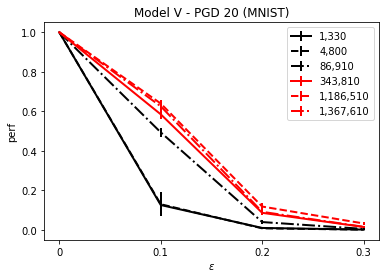}
    \caption{Standard accuracy ($\epsilon=0$) and robust accuracy ($\epsilon > 0$) of models over MNIST dataset for the unnormalized setting corresponding to Fig.~\ref{fig:NORMALIZED_SUM}.}
    \label{fig:mnist_normalized}
\end{figure}

\begin{figure}[t]
    \centering
    \includegraphics[width=.45\textwidth]{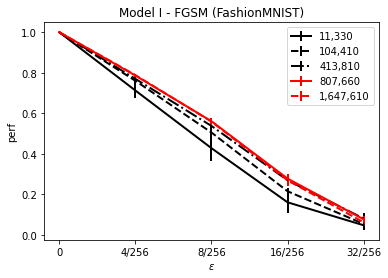}
    \includegraphics[width=.45\textwidth]{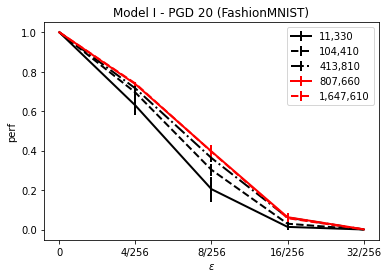} \\
    \includegraphics[width=.45\textwidth]{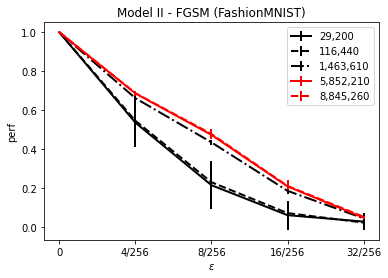}
    \includegraphics[width=.45\textwidth]{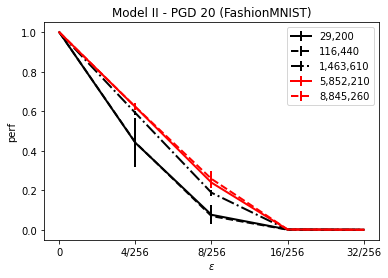} \\
    \includegraphics[width=.45\textwidth]{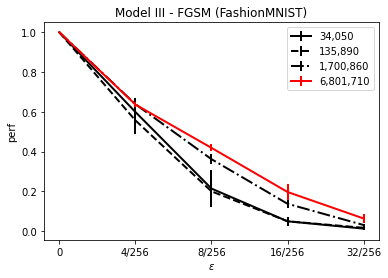}
    \includegraphics[width=.45\textwidth]{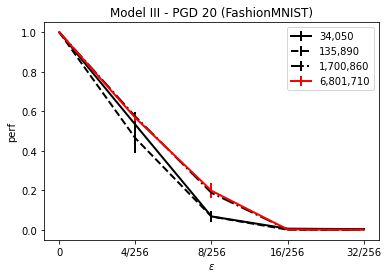} \\
    \includegraphics[width=.45\textwidth]{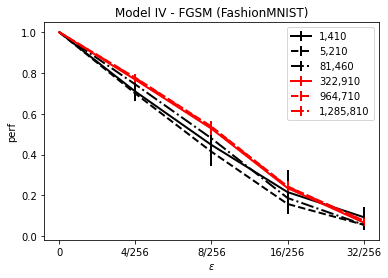}
    \includegraphics[width=.45\textwidth]{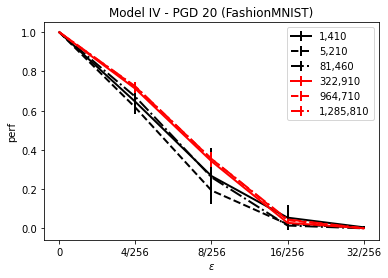} \\
    \includegraphics[width=.45\textwidth]{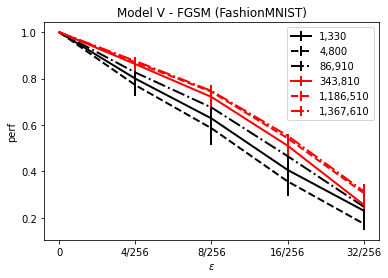}
    \includegraphics[width=.45\textwidth]{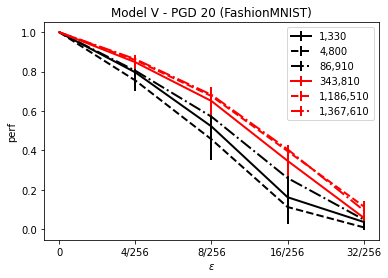}
    \caption{Standard accuracy ($\epsilon=0$) and robust accuracy ($\epsilon > 0$) of models over FashionMNIST dataset for the unnormalized setting corresponding to Fig.~\ref{fig:NORMALIZED_SUM}.}
    \label{fig:fmnist_normalized}
\end{figure}

\begin{figure}[t]
    \centering
    \includegraphics[width=.45\textwidth]{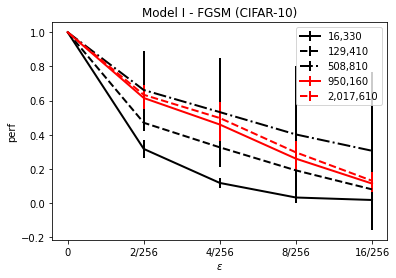}
    \includegraphics[width=.45\textwidth]{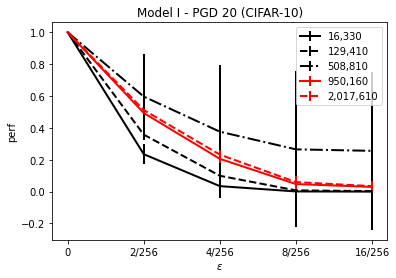} \\
    \includegraphics[width=.45\textwidth]{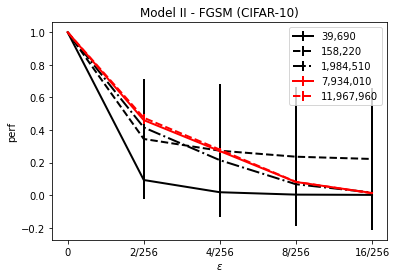}
    \includegraphics[width=.45\textwidth]{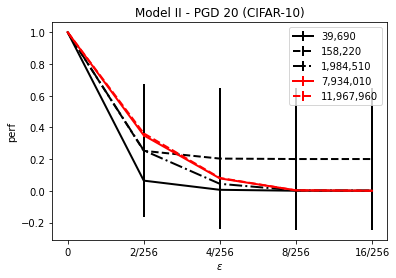} \\
    \includegraphics[width=.45\textwidth]{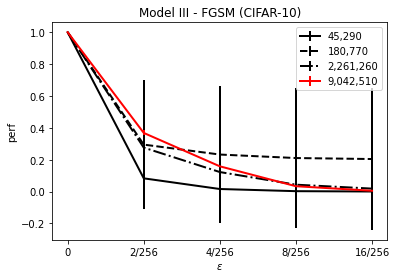}
    \includegraphics[width=.45\textwidth]{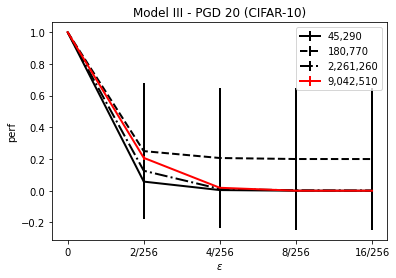} \\
    \includegraphics[width=.45\textwidth]{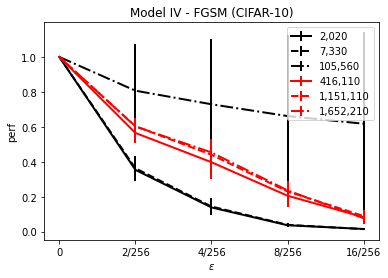}
    \includegraphics[width=.45\textwidth]{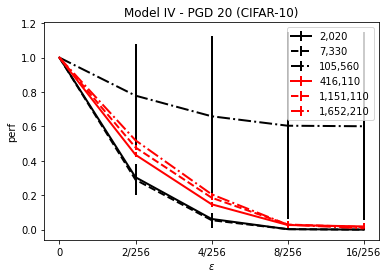} \\
    \includegraphics[width=.45\textwidth]{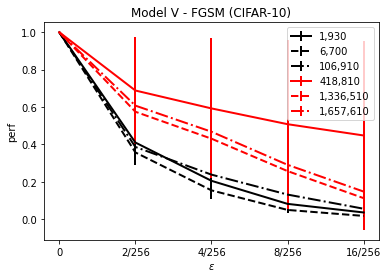}
    \includegraphics[width=.45\textwidth]{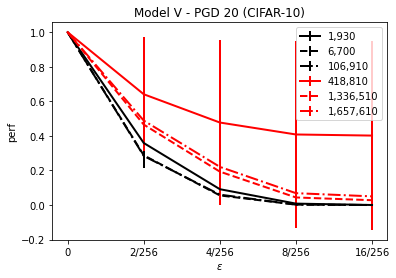}
    \caption{Standard accuracy ($\epsilon=0$) and robust accuracy ($\epsilon > 0$) of models over CIFAR-10 dataset for the unnormalized setting corresponding to Fig.~\ref{fig:NORMALIZED_SUM}.}
    \label{fig:cifar_normalized}
\end{figure}

\end{document}